\documentclass[runningheads]{llncs}

\usepackage{eccvabbrv}

\usepackage{graphicx}
\usepackage{booktabs}

\usepackage[accsupp]{axessibility}  
\usepackage{bm}
\def\eg{\emph{e.g.,}}

\def\ie{\emph{i.e.,}}

\newcommand{\model}{\epsilon_\theta}
\newcommand{\z}[1]{z}
\newcommand{\encoder}{\mathcal{E}}
\newcommand{\conditioner}{\encoder_\theta}
\newcommand{\ours}{\textup{DragAnything}\xspace}

\usepackage[pagebackref,breaklinks,colorlinks]{hyperref}

\usepackage{orcidlink}

\begin{document}

\title{\ours: Motion Control for Anything using Entity Representation}


\titlerunning{Abbreviated paper title}

\author{Weijia Wu\inst{1}\inst{2}\inst{3}
\and
Zhuang Li\inst{1} \and
Yuchao Gu\inst{3} \and
Rui Zhao\inst{3} \and
Yefei He\inst{2}  \and
David Junhao Zhang\inst{3}
\and Mike Zheng Shou\inst{3}$^\dagger$
\and
Yan Li\inst{1} \and
Tingting Gao\inst{1}
\and Di Zhang\inst{1}
}

\authorrunning{Wu et al.}

\institute{Kuaishou Technology \and Zhejiang University \and
Show Lab, National University of Singapore}

\maketitle

\begin{abstract}
  %
  We introduce \ours, which utilizes a entity representation to achieve motion control for any object in controllable video generation.
  Comparison to existing motion control methods, \ours offers several advantages. 
  Firstly, trajectory-based is more user-friendly for interaction, when acquiring other guidance signals (\eg{} masks, depth maps) is labor-intensive. 
  Users only need to draw a line~(trajectory) during interaction.
  Secondly, our entity representation serves as an open-domain embedding capable of representing any object, enabling the control of motion for diverse entities, including background.
  Lastly, our entity representation allows simultaneous and distinct motion control for multiple objects.
  Extensive experiments demonstrate that our \ours achieves state-of-the-art performance for FVD, FID, and User Study, particularly in terms of object motion control, where our method surpasses the previous methods (\eg{} DragNUWA) by $26\%$ in human voting.

  The project website is at:  
   \href{https://weijiawu.github.io/draganything_page/}{\color{blue}{\tt DragAnything}}.

  \keywords{Motion Control \and Controllable Video Generation \and Diffusion Model}
\end{abstract}

\begin{figure*}[t]
	\includegraphics[width=0.99\linewidth]{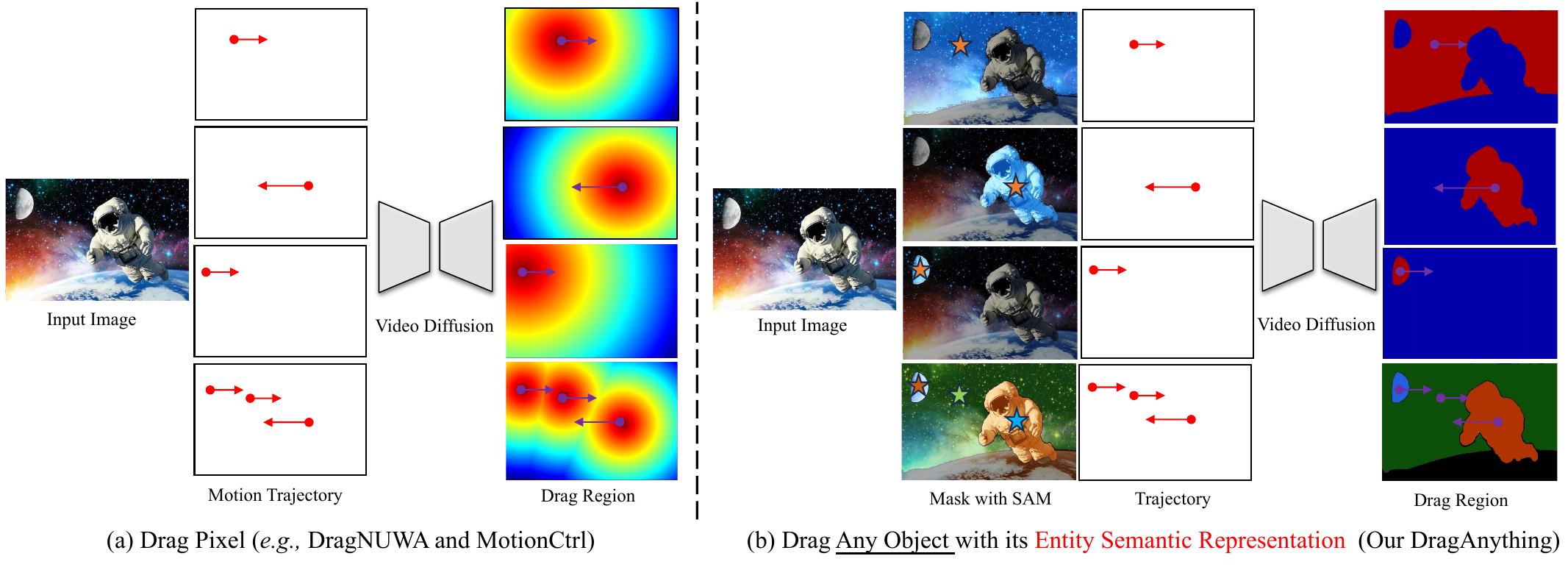}
	\vspace{-0.2cm}
	\caption{\textbf{Comparison with Previous Works.} (a) Previous works (Motionctrl~\cite{wang2023motionctrl}, DragNUWA~\cite{yin2023dragnuwa}) achieved motion control by dragging pixel points or pixel regions. (b) \ours enables more precise entity-level motion control by manipulating the corresponding entity representation.
    }
    \vspace{-0.45cm}
\label{motivation}
\end{figure*}

\section{Introduction}
\label{sec:intro}
Recently, there have been significant advancements in video generation, 
with notable works such as Imagen Video~\cite{ho2022imagen}, Gen-2 ~\cite{esser2023structure}, PikaLab~\cite{pikalab}, SVD~\cite{blattmann2023stable}, and SORA~\cite{SORA} garnering considerable attention from the community.
%
%
However, the pursuit of controllable video generation has encountered relatively slower progress, notwithstanding its pivotal significance.
Unlike controllable static image generation~\cite{zhang2023adding,rombach2022high,ramesh2022hierarchical}, controllable video generation poses a more intricate challenge, demanding not only spatial content manipulation but also precise temporal motion control.

Recently, trajectory-based motion control~\cite{hao2018controllable,ardino2021click,wang2023motionctrl,yin2023dragnuwa} has been proven to be a user-friendly and efficient solution for controllable video generation. 
Compared to other guidance signals such as masks or depth maps, drawing a trajectory provides a simple and flexible approach.
Early trajectory-based~\cite{hao2018controllable,ardino2021click,blattmann2021ipoke,blattmann2021understanding} works utilized optical flow or recurrent neural networks to control the motion of objects in controllable video generation.
As one of the representative works, DragNUWA~\cite{yin2023dragnuwa} encodes sparse strokes into dense flow space, which is then used as a guidance signal for controlling the motion of objects.
Similarly, MotionCtrl~\cite{wang2023motionctrl} directly encodes the trajectory coordinates of each object into a vector map, using this vector map as a condition to control the motion of the object.
These works have made significant contributions to the controllable video generation. 
However, an important question has been overlooked: \textit{Can a single point on the target truly represent the target?}

\begin{figure*}[t]
	\includegraphics[width=0.99\linewidth]{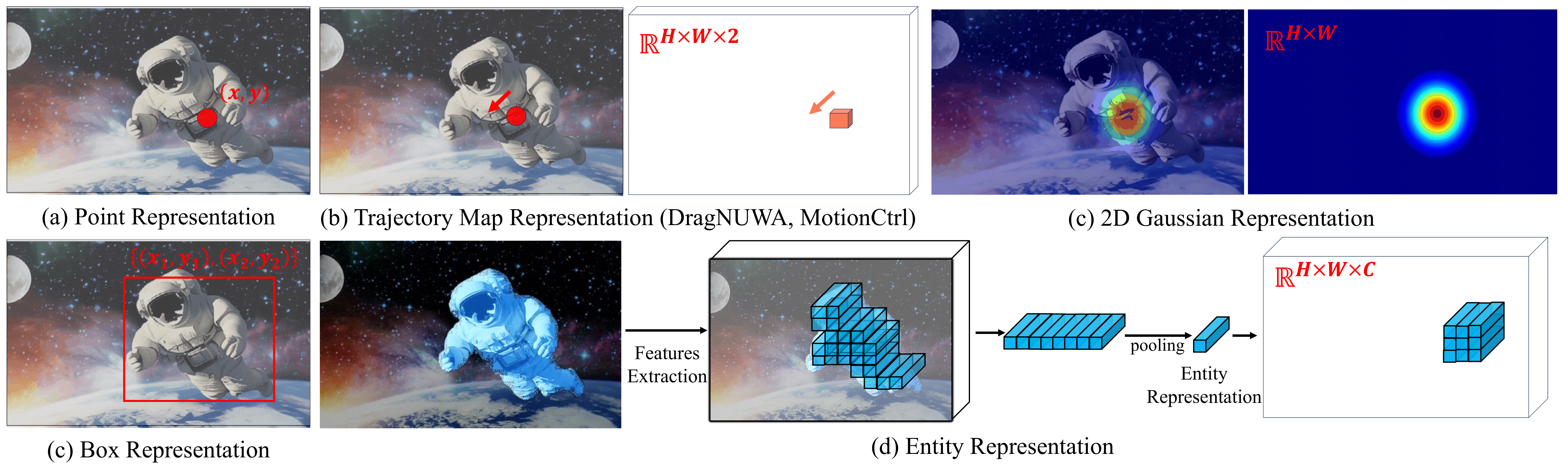}
	\vspace{-0.2cm}
	\caption{\textbf{Comparison for Different Representation Modeling.} 
     (a) Point representation: using a coordinate point $(x,y)$ to represent an entity.  (b) Trajectory Map: using a trajectory vector map to represent the trajectory of the entity. (c) 2D gaussian: using a 2D Gaussian map to represent an entity. (c) Box representation: using a bounding box to represent an entity. (d) Entity representation: extracting the latent diffusion feature of the entity to characterize it.
    }
    \vspace{-0.25cm}
\label{comparison1}
\end{figure*}

Certainly, a single pixel point cannot represent an entire object, as shown in Figure~\ref{comparison1} (a)-(b).
Thus, dragging a single pixel point may not precisely control the object it corresponds to.
As shown in Figure~\ref{motivation}, 
given the trajectory of a pixel on a star of starry sky, the model may not distinguish between controlling the motion of the star or that of the entire starry sky; it merely drags the associated pixel area.
Indeed, resolving this issue requires clarifying two concepts: 1) \textbf{What entity.} Identifying the specific area or entity to be dragged. 
2) \textbf{How to drag.} How to achieve dragging only the selected area, meaning separating the background from the foreground that needs to be dragged.
For the first challenge, interactive segmentation~\cite{kirillov2023segment,wang2023seggpt} is an efficient solution.
For instance, in the initial frame, employing SAM~\cite{kirillov2023segment} allows us to conveniently select the region we want to control.
In comparison, the second technical issue poses a greater challenge. 
To address this, this paper proposes a novel Entity Representation to achieve precise motion control for any entity in the video.

Some works~\cite{chen2023anydoor,gu2023videoswap,tang2024emergent} has already demonstrated the effectiveness of using latent features to represent corresponding objects.
Anydoor~\cite{chen2023anydoor} utilizes features from Dino v2~\cite{oquab2023dinov2} to handle object customization, while VideoSwap~\cite{gu2023videoswap} and DIFT~\cite{tang2024emergent} employ features from the diffusion model~\cite{rombach2022high} to address video editing tasks.
Inspired by these works, we present \ours , which utilize the latent feature of the diffusion model to represent each entity.
As shown in Figure~\ref{comparison1} (d), based on the coordinate indices of the entity mask, we can extract the corresponding semantic features from the diffusion feature of the first frame. 
We then use these features to represent the entity, achieving entity-level motion control by manipulating the spatial position of the corresponding latent feature.

In our work, \ours employs SVD~\cite{blattmann2023stable} as the foundational model. 
Training \ours requires video data along with the motion trajectory points and the entity mask of the first frame. 
To obtain the required data and annotations, we utilize the video segmentation benchmark~\cite{miao2022large} to train \ours.
The mask of each entity in the first frame is used to extract the central coordinate of that entity, and then CoTrack~\cite{karaev2023cotracker} is utilized to predict the motion trajectory of the point as the entity motion trajectory.

Our main contributions are summarized as follows:
\begin{itemize}
\itemsep -0cm

    \item
    New insights for trajectory-based controllable generation that reveal the differences between pixel-level motion and entity-level motion.
    
    \item
    Different from the drag pixel paradigm,
    we present \ours, which can achieve true entity-level motion control with the entity representation.

    \item
    \ours achieves SOTA performance for FVD, FID, and User Study, surpassing the previous method by $26\%$ in human voting for motion control.
    \ours supports interactive motion control for anything in context, including background (\eg{} \texttt{sky}), as shown in Figure~\ref{vis} and Figure~\ref{vis2}.

\end{itemize}


\section{Related Works}

\subsection{Image and Video Generation}
Recently, image generation~\cite{rombach2022high,ramesh2022hierarchical,wu2023paragraph,gu2024mix,wu2023diffumask,he2024ptqd,he2023efficientdm} has attracted considerable attention.
Some notable works, such as Stable Diffusion~\cite{rombach2022high} of Stability AI, DALL-E2~\cite{ramesh2022hierarchical} of OpenAI, Imagen~\cite{saharia2022photorealistic} of Google, RAPHAEL~\cite{xue2023raphael} of SenseTime, and Emu~\cite{dai2023emu} of Meta, have made significant strides, contributions, and impact in the domain of image generation tasks.
Controllable image generation has also seen significant development and progress, exemplified by ControlNet~\cite{zhang2023adding}.
By utilizing guidance information such as Canny edges, Hough lines, user scribbles, human key points, segmentation maps, precise image generation can be achieved.

In contrast, progress~\cite{xing2023survey,wu2023tune,wang2023lavie,chen2023videocrafter1,zhou2022magicvideo,zhang2023show} in the field of video generation is still relatively early-stage.
Video diffusion models~\cite{ho2022video} was first introduced using a 3D U-Net diffusion model architecture to predict and generate a sequence of videos.
Imagen Video~\cite{ho2022imagen} proposed a cascaded diffusion video model for high-definition video generation, and attempt to transfer the text-to-image setting to video generation.
Show-1~\cite{zhang2023show} directly implements a temporal diffusion model in pixel space, and utilizes inpainting and super-resolution for high-resolution synthesis.
Video LDM~\cite{blattmann2023align} marks the first application of the LDM paradigm to high-resolution video generation, introducing a temporal dimension to the latent space diffusion model.
I2vgen-xl~\cite{zhang2023i2vgen} introduces a cascaded network that improves model performance by separating these two factors and ensures data alignment by incorporating static images as essential guidance.
Apart from academic research, the industry has also produced numerous notable works, including Gen-2 ~\cite{esser2023structure}, PikaLab~\cite{pikalab}, and SORA~\cite{SORA}.
However, compared to the general video generation efforts, the development of controllable video generation still has room for improvement.
In our work, we aim to advance the field of trajectory-based video generation.

\subsection{Controllable Video Generation}
There have been some efforts~\cite{zhang2023controlvideo,ma2023follow,chen2023motion,guo2023sparsectrl,ma2023trailblazer,zhang2024moonshot} focused on controllable video generation, such as AnimateDiff~\cite{guo2023animatediff}, Control-A-Video~\cite{chen2023control}, Emu Video~\cite{girdhar2023emu}, and Motiondirector~\cite{zhao2023motiondirector}.
Control-A-Video~\cite{chen2023control} attempts to generate videos conditioned on a sequence of control signals, such as edge or depth maps, with two motion-adaptive noise initialization strategies.
Follow Your Pose~\cite{ma2023follow} propose a two-stage training
scheme that can utilize image pose pair and pose-free video to obtain the pose-controllable character videos. 
ControlVideo~\cite{zhang2023controlvideo} design a training-free framework to enable controllable text-to-video generation with structural consistency.
These works all focus on video generation tasks guided by dense guidance signals (such as masks, human poses, depth).
However, obtaining dense guidance signals in real-world applications is challenging and not user-friendly.
By comparison, using a trajectory-based approach for drag seems more feasible.

Early trajectory-based works~\cite{hao2018controllable,ardino2021click,blattmann2021ipoke,blattmann2021understanding} often utilized optical flow or recurrent neural networks to achieve motion control.
TrailBlazer~\cite{ma2023trailblazer} focuses on enhancing controllability in video synthesis by employing bounding boxes to guide the motion of subject.
DragNUWA~\cite{yin2023dragnuwa} encodes sparse strokes into a dense flow space, subsequently employing this as a guidance signal to control the motion of objects.
Similarly, MotionCtrl~\cite{wang2023motionctrl} directly encodes the trajectory coordinates of each object into a vector map, using it as a condition to control the object's motion.
These works can be categorized into two paradigms: Trajectory Map (point) and box representation.
The box representation~(\eg{} TrailBlazer~\cite{ma2023trailblazer}) only handle instance-level objects and cannot accommodate backgrounds such as starry skies. 
Existing Trajectory Map Representation ~(\eg{} DragNUWA, MotionCtrl) methods are quite crude, as they do not consider the semantic aspects of entities.
In other words, a single point cannot adequately represent an entity.
In our paper, we introduce \ours, which can achieve true entity-level motion control using the proposed entity representation.



\section{Methodology}

%
%

\subsection{Task Formulation and Motivation}
\label{Taskmotivation}

\subsubsection{Task Formulation.} The trajectory-based video generation task requires the model to synthesize videos based on given motion trajectories.
Given a point trajectories ${(x_1, y_1), (x_2, y_2), \dots, (x_L, y_L)}$, where $L$ denotes the video length, a conditional denoising autoencoder $\model(z , c)$ is utilized to generate videos that correspond to the motion trajectory.
The guidance signal $c$ in our paper encompasses three types of information: trajectory points, the first frame of the video, and the entity mask of the first frame.

 \subsubsection{Motivation.} Recently, some trajectory-based works, such as DragNUWA~\cite{yin2023dragnuwa} and MotionCtrl~\cite{wang2023motionctrl} have explored using trajectory points to control the motion of objects in video generation.
 These approaches typically directly manipulate corresponding pixels or pixel areas using the provided trajectory coordinates or their derivatives. 
 However, they overlook a crucial issue:
 As shown in Figure~\ref{motivation} and Figure~\ref{comparison1}, \textit{\textbf{the provided trajectory points may not fully represent the entity we intend to control}}.
 Therefore, dragging these points may not necessarily correctly control the motion of the object.

\begin{figure*}[t]
    \begin{center}
	\includegraphics[width=0.99\linewidth]{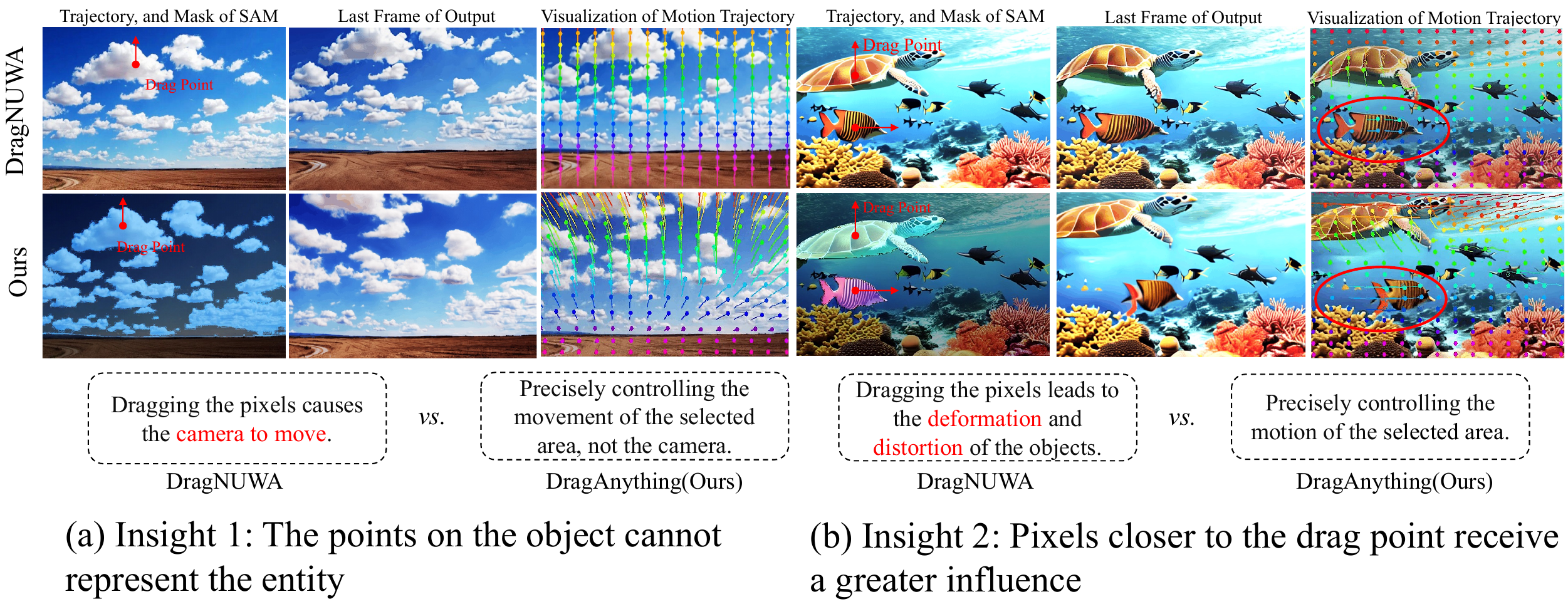}
    \end{center}
	\vspace{-0.2cm}
	\caption{\textbf{Toy experiment for the motivation of Entity Representation.} Existing methods (DragNUWA~\cite{yin2023dragnuwa} and MotionCtrl~\cite{wang2023motionctrl}) directly drag pixels, which cannot precisely control object targets, whereas our method employs entity representation to achieve precise control. 
    %
    }
    \vspace{-0.25cm}
\label{motivation_method}
\end{figure*}

 To validate our hypothesis, \textit{i.e.,} that simply dragging pixels or pixel regions cannot effectively control object motion, we designed a toy experiment to confirm.
 As shown in Figure~\ref{motivation_method}, we employed a classic point tracker, \ie{} Co-Tracker~\cite{karaev2023cotracker}, to track every pixel in the synthesized video and observe their trajectory changes. 
From the change in pixel motion, we gain two new insights:

\paragraph{Insight 1: The trajectory points on the object cannot represent the entity.}
(Figure~\ref{motivation_method} (a)). 
%
%
From the pixel motion trajectories of DragUNWA, it is evident that dragging a pixel point of the \texttt{cloud} does not cause the \texttt{cloud} to move; instead, it results in the camera moving up.
This indicates that the model cannot perceive our intention to control the cloud, implying that a single point cannot represent the cloud.
Therefore, we pondered whether there exists a more direct and effective representation that can precisely control the region we intend to manipulate (the selected area).

\paragraph{Insight 2: For the trajectory point representation paradigm (Figure~\ref{comparison1} (a)-(c)), pixels closer to the drag point receive a greater influence, resulting in larger motions (Figure~\ref{motivation_method} (b)).} 
By comparison, we observe that in the videos synthesized by DragNUWA, pixels closer to the drag point exhibit larger motion.
However, what we expect is for the object to move as a whole according to the provided trajectory, rather than individual pixel motion.

Based on the above two new insights and observations, we present a novel Entity Representation, which extracts latent features of the object we want to control as its representation.
As shown in Figure~\ref{motivation_method}, visualization of the corresponding motion trajectories shows that our method can achieve more precise entity-level motion control.
For example, Figure~\ref{motivation_method} (b) shows that our method can precisely control the motion of \texttt{seagulls} and \texttt{fish}, while DragNUWA only drags the movement of corresponding pixel regions, resulting in abnormal deformation of the appearance.
%
%

\begin{figure*}[t]
	\includegraphics[width=0.99\linewidth]{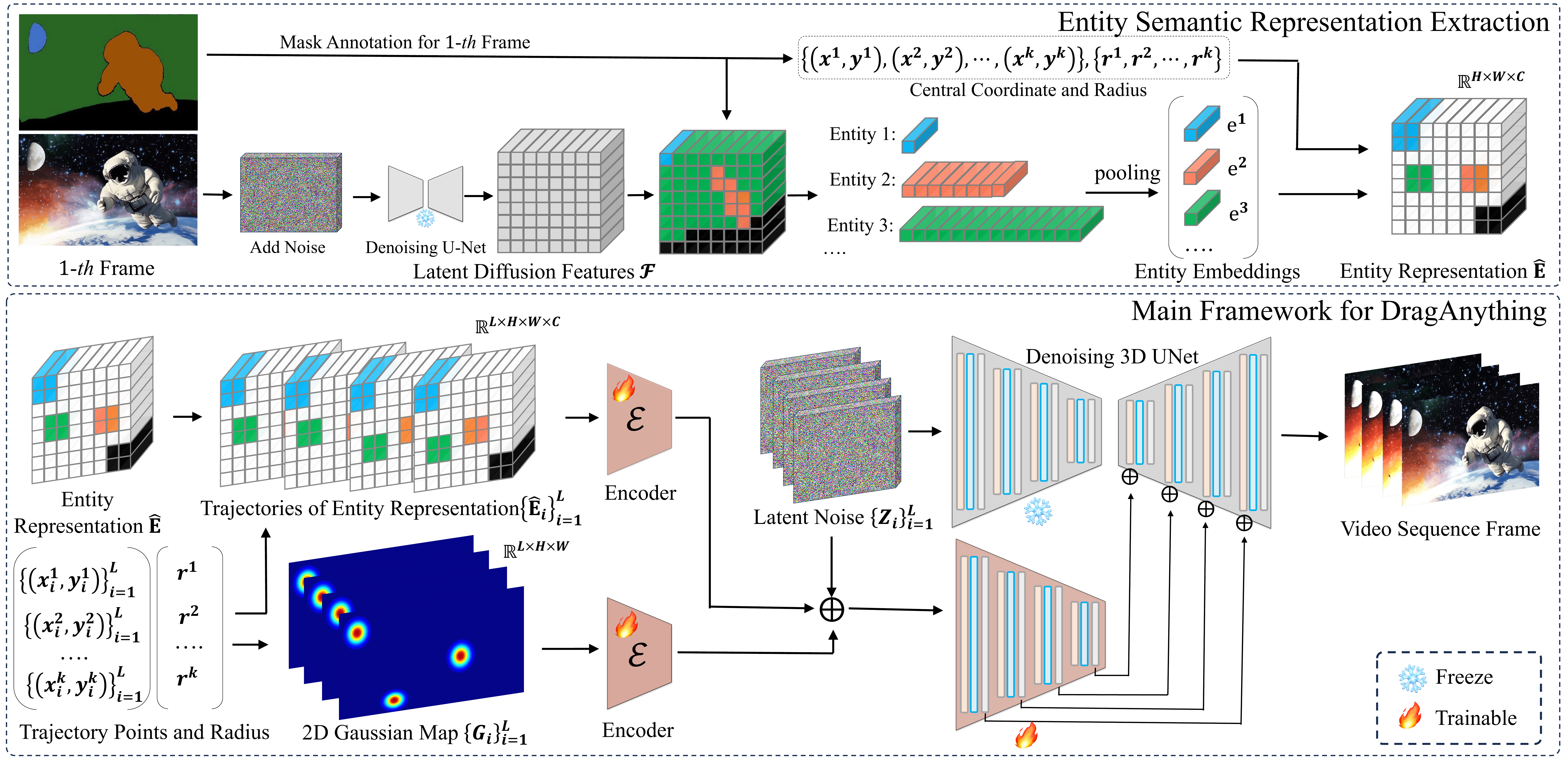}
	\vspace{-0.2cm}
	\caption{\textbf{\ours Framework.} The architecture includes two parts: 
    1) Entity Semantic Representation Extraction. Latent features from the Diffusion Model are extracted based on entity mask indices to serve as corresponding entity representations. 
    2) Main Framework for \ours. Utilizing the corresponding entity representations and 2D Gaussian representations to control the motion of entities.
    }
    \vspace{-0.25cm}
\label{Framework}
\end{figure*}

\subsection{Architecture}
\label{Architecture}
Following SVD \cite{blattmann2023stable}, our base architecture mainly consists of three components: a denoising diffusion model (3D U-Net~\cite{unet}) to learn the denoising process for space and time efficiency, an encoder  and a decoder, to encode videos into the latent space and reconstruct the denoised latent features back into videos.
Inspired by Controlnet~\cite{zhang2023adding}, we adopt a 3D Unet to encode our guidance signal, which is then applied to the decoder blocks of the denoising 3D Unet of SVD, as shown in Figure~\ref{Framework}.
Different from the previous works, we designed an entity representation extraction mechanism and combined it with 2D Gaussian representation to form the final effective representation.
Then we can achieve entity-level controllable generation with the representation.

\subsection{Entity Semantic Representation Extraction}
\label{Entity}
The conditional signal of our method requires  gaussian representation~(\S\ref{gaussian}) and the corresponding entity representation~(\S\ref{Entity}).
In this section, we describe how to extract these representations from the first frame image.

\subsubsection{Entity Representation Extraction.}
\label{Entity1}
Given the first frame image $\bm{\mathrm{I}} \in \mathbb{R}^{H\times W \times 3}$ with the corresponding entity mask $\bm{\mathrm{M}}$,
we first obtain the latent noise $\bm{x}$ of the image through diffusion inversion (diffusion forward process)~\cite{ho2020denoising,wu2024datasetdm,tang2024emergent}, which is not trainable and is based on a fixed Markov chain that gradually adds Gaussian noise to the image.
Then, a denoising U-Net $\epsilon_{\theta}$ is used to extract the corresponding latent diffusion features $\mathcal{F} \in \mathbb{R}^{H\times W \times C}$ as follows: 
\begin{align}
  \mathcal{F} = \epsilon_{\theta}(\bm{x}_t, t),
\end{align}
where $t$ represents the $t$-th time step. 
Previous works~\cite{tang2024emergent,gu2023videoswap,wu2024datasetdm} has already demonstrated the effectiveness of a single forward pass for representation extraction, and extracting features from just one step has two advantages: faster inference speed and better performance.
With the diffusion features $\mathcal{F}$, the corresponding entity embeddings can be obtained by indexing the corresponding coordinates from the entity mask.
For convenience, average pooling is used to process the corresponding entity embeddings to obtain the final embedding $\{e_1, e_2, ..., e_k\}$, where $k$ denotes the number of entity and each of them has a channel size of $C$.

To associate these entity embeddings with the corresponding trajectory points, we directly initialize a zero matrix $\bm{\mathrm{E}} \in \mathbb{R}^{H\times W \times C}$ and then insert the entity embeddings based on the trajectory sequence points, as shown in Figure~\ref{groundtruth}.
During the training process, we use the entity mask of the first frame to extract the center coordinates $\{(x^1,y^1), (x^2,y^2), ..., (x^k,y^k)\}$ of the entity as the starting point for each trajectory sequence point.
With these center coordinate indices, the final entity representation $\bm{\mathrm{\hat{E}}}$ can be obtained by inserting the entity embeddings into the corresponding zero matrix $\bm{\mathrm{E}}$ (Deatils see Section~\ref{label}).

%
With the center coordinates $\{(x^1,y^1), (x^2,y^2), ..., (x^k,y^k)\}$ of the entity in the first frame, we use Co-Tracker~\cite{karaev2023cotracker} to track these points and obtain the corresponding motion trajectories $\{\{(x^1_i,y^1_i)\}_{i=1}^L,\{(x^2_i,y^2_i)\}_{i=1}^L,...,\{(x^k_i,y^k_i)\}_{i=1}^L\}$, where $L$ is the length of video.
Then we can obtain the corresponding entity representation $\{\bm{\mathrm{\hat{E}}_i}\}_{i=1}^L$ for each frame.

\subsubsection{2D Gaussian Representation Extraction.}
\label{gaussian}
Pixels closer to the center of the entity are typically more important.
We aim to make the proposed entity representation focus more on the central region, while reducing the weight of edge pixels.
The 2D Gaussian Representation can effectively enhance this aspect, with pixels closer to the center carrying greater weight, as illustrated in Figure~\ref{comparison1} (c).
With the point trajectories $\{\{(x^1_i,y^1_i)\}_{i=1}^L,\{(x^2_i,y^2_i)\}_{i=1}^L,...,\{(x^k_i,y^k_i)\}_{i=1}^L\}$ and $\{r^1,...,r^k\}$, we can obtain the corresponding 2D Gaussian Distribution Representation trajectory sequences $\{\bm{\mathrm{G}_i}\}_{i=1}^L$, as illustrated in Figure~\ref{groundtruth}.
Then, after processing with a encoder $\encoder$ (see Section~\ref{encoder}), we merge it with the entity representation to achieve enhanced focus on the central region performance, as shown in Figure~\ref{Framework}.

\subsubsection{Encoder for Entity Representation and 2D Gaussian Map.}
\label{encoder}
As shown in Figure~\ref{Framework}, 
the encoder, denoted as $\encoder$, is utilized to encode the entity representation and 2D Gaussian map into the latent feature space.
In this encoder, we utilized four blocks of convolution to process the corresponding input guidance signal, where each block consists of two convolutional layers and one SiLU activation function.
Each block downsamples the input feature resolution by a factor of 2, resulting in a final output resolution of $1/8$.
The encoder structure for processing the entity and gaussian representation is the same, with the only difference being the number of channels in the first block, which varies when the channels for the two representations are different.
After passing through the encoder, we follow ControlNet~\cite{zhang2023adding} by adding the latent features of Entity Representation and 2D Gaussian Map Representation with the corresponding latent noise of the video:
\begin{align}
  \{\bm{\mathrm{R}_i}\}_{i=1}^L = \encoder(\{\bm{\mathrm{\hat{E}}_i}\}_{i=1}^L) + \encoder(\{\bm{\mathrm{G}_i}\}_{i=1}^L) + \{\bm{\mathrm{Z}_i}\}_{i=1}^L  ,
  \label{equ2}
\end{align}
where $\bm{\mathrm{Z}_i}$ denotes the latent noise of $i$-th frame.
Then the feature $\{\bm{\mathrm{R}_i}\}_{i=1}^L$ is inputted into the encoder of the denoising 3D Unet to obtain four features with different resolutions, which serve as latent condition signals. 
The four features are added to the feature of the denoising 3D Unet of the foundation model.

\subsection{Training and Inference}
\label{Training}
\label{label}
\begin{figure*}[t]
	\includegraphics[width=0.99\linewidth]{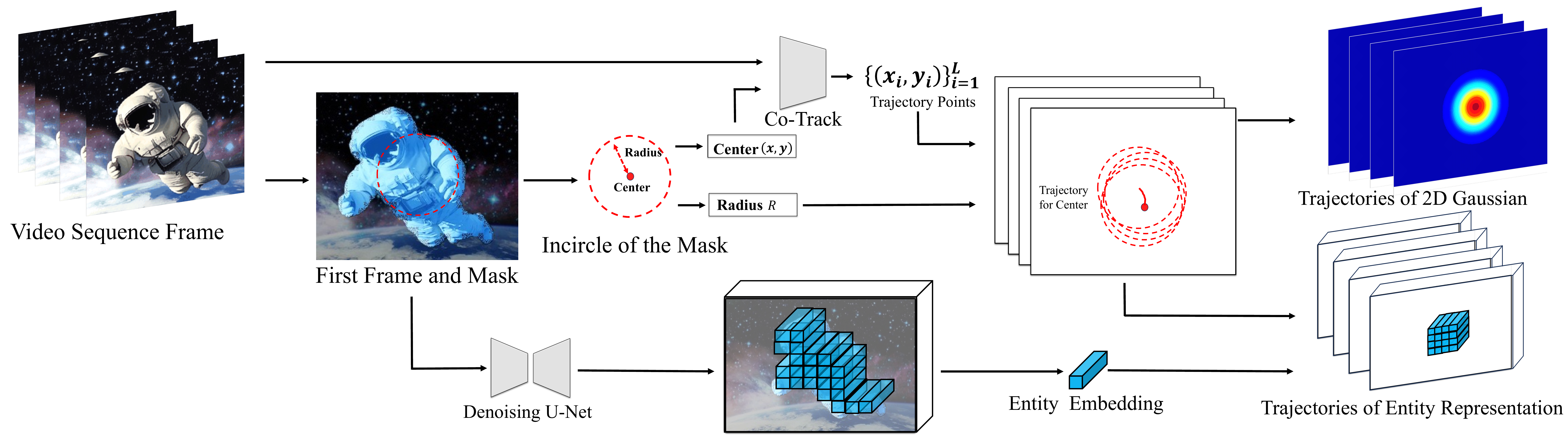}
	\vspace{-0.2cm}
	\caption{\textbf{ Illustration of ground truth generation procedure.} During the training process, we generate ground truth labels from video segmentation datasets that have entity-level annotations.
    }
    \vspace{-0.25cm}
\label{groundtruth}
\end{figure*}

\subsubsection{Ground Truth Label Generation.}
\label{groundtruth1}
During the training process, we need to generate corresponding Trajectories of Entity Representation and 2D Gaussian, as shown in Figure~\ref{groundtruth}.
First, for each entity, we calculate its incircle circle using its corresponding mask, obtaining its center coordinates $(x,y)$ and radius $r$.
Then we use Co-Tracker~\cite{karaev2023cotracker} to obtain its corresponding trajectory of the center $\{(x_i,y_i)\}_{i=1}^L$, serving as the representative motion trajectory of that entity.
With these trajectory points and radius, we can calculate the corresponding Gaussian distribution value~\cite{cao2017realtime} at each frame.
For entity representation, we insert the corresponding entity embedding into the circle centered at $(x,y)$ coordinates with a radius of $r$.
Finally, we obtain the corresponding trajectories of Entity Representation and 2D Gaussian for training our model.

\subsubsection{Loss Function.} In video generation tasks, Mean Squared Error (MSE) is commonly used to optimize the model.
Given the corresponding entity representation $\bm{\mathrm{\hat{E}}}$ and 2D Gaussian representation $\bm{\mathrm{G}}$, the objective can be simplified to:
\begin{align}
    \mathcal{L}_\theta=\sum_{i=1}^{L} \bm{\mathrm{M}}\left|\left|\epsilon-\epsilon_\theta\left(\bm{x}_{t,i},\conditioner(\bm{\mathrm{\hat{E}}}_i),\conditioner(\bm{\mathrm{G}}_i)\right)\right|\right|_2^2\, ,
\end{align}
where $\conditioner$ denotes the encoder for entity and 2d gaussian representations.
$\bm{\mathrm{M}}$ is the mask for entities of images at each frame.
The optimization objective of the model is to control the motion of the target object.
For other objects or the background, we do not want to affect the generation quality.
Therefore, we use a mask $\bm{\mathrm{M}}$ to constrain the MSE loss to only backpropagate through the areas we want to optimize.

\subsubsection{Inference of User-Trajectory Interaction.}
\ours is user-friendly.
During inference, the user only needs to click to select the region they want to control with SAM~\cite{kirillov2023segment},
and then drag any pixel within the region to form a reasonable trajectory.
Our \ours can then generate a video that corresponds to the desired motion.

\section{Experiments}

\subsection{Experiment Settings}

\textbf{Implementation Details.} 
Our \ours is based on the Stable Video Diffusion (SVD)~\cite{blattmann2023stable} architecture and weights, which were trained to generate $25$ frames at a resolution of $320\times576$.
All the experiments are conducted on PyTorch with Tesla A100 GPUs.
AdamW~\cite{loshchilov2017decoupled} as the optimizer for total $100k$ training steps with the learning rate of 1e\mbox{-}5. 
%
%

\begin{figure*}[t]
    \begin{center}
	\includegraphics[width=0.99\linewidth]{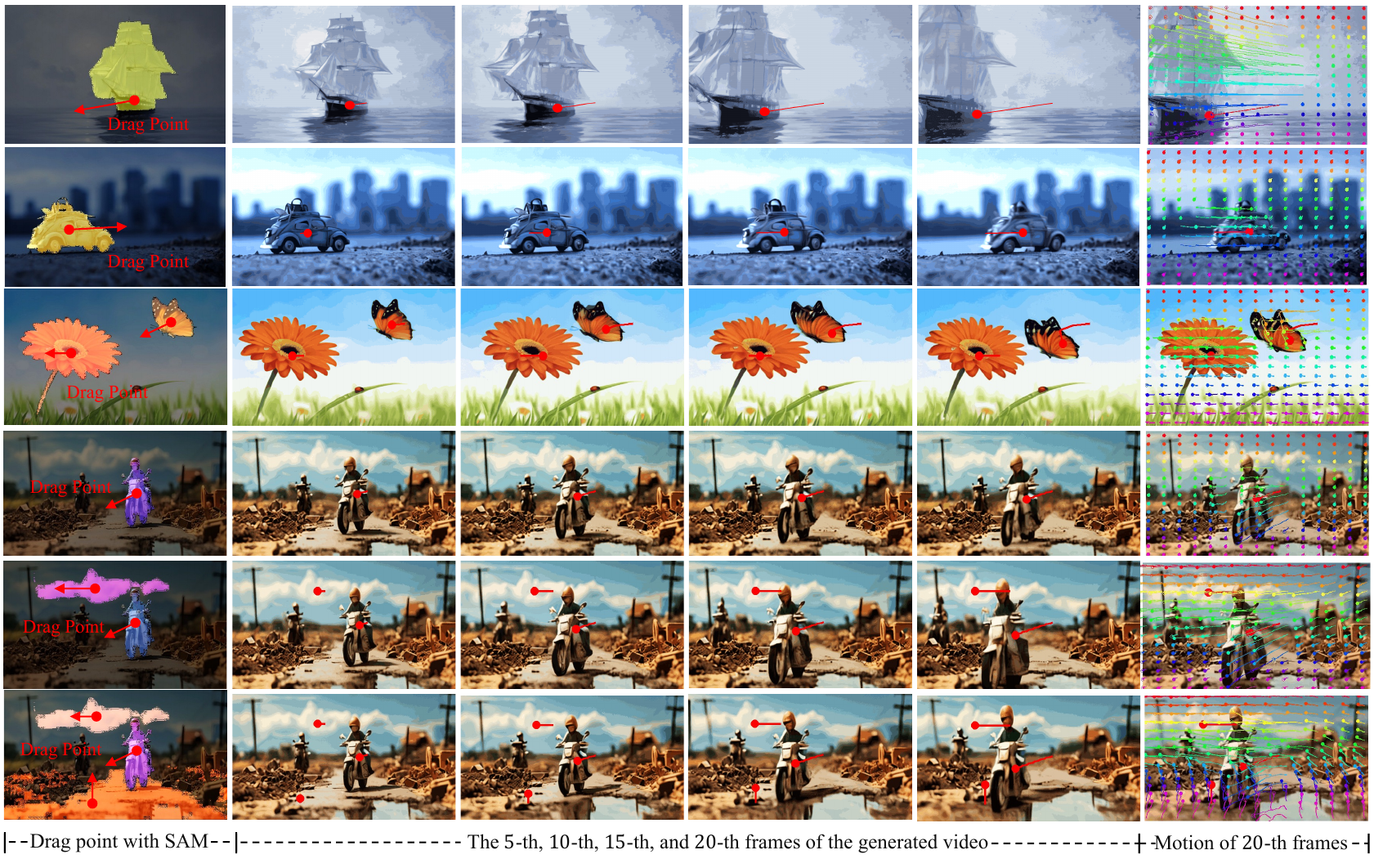}
    \end{center}
	\vspace{-0.6cm}
	\caption{\textbf{Visualization for \ours.} The proposed \ours can accurately control the motion of objects at the entity level, producing high-quality videos. 
    The visualization for the pixel motion of $20$-th frame is obatined by Co-Track~\cite{karaev2023cotracker}.
    }
    \vspace{-0.25cm}
\label{vis}
\end{figure*}

\textbf{Evaluation Metrics.}
To comprehensively evaluate our approach, we conducted evaluations from both human assessment and automatic script metrics perspectives. 
Following MotionControl~\cite{wang2023motionctrl}, we employed two types of automatic script metrics:
1) \textit{Evaluation of video quality}: We utilized Frechet Inception Distance (FID)~\cite{fid} and Frechet Video Distance (FVD)~\cite{fvd} to assess visual quality and temporal coherence.
2) \textit{Assessment of object motion control performance}: The Euclidean distance between the predicted and ground truth object trajectories (ObjMC) was employed to evaluate object motion control.
In addition, for the user study, considering video aesthetics, we collected and annotate $30$ images from Google Image along with their corresponding point trajectories and the corresponding mask.
Three professional evaluators are required to vote on the synthesized videos from two aspects: video quality and motion matching.
The videos of Figure~\ref{vis} and Figure~\ref{vis2} are sampled from these $30$ cases.

\textbf{Datasets.} Evaluation for the trajectory-guided video generation task requires the motion trajectory of each video in the test set as input.
To obtain such annotated data, we adopted the VIPSeg~\cite{miao2022large} validation set as our test set. 
We utilized the instance mask of each object in the first frame of the video, extracted its central coordinate, and employed Co-Tracker~\cite{karaev2023cotracker} to track this point and obtain the corresponding motion trajectory as the ground truth for metric evaluation.
As FVD requires videos to have the same resolution and length, we resized the VIPSeg \texttt{val} dataset to a resolution of $256 \times 256$ and a length of 14 frames for evaluation.
Correspondingly, we also utilized the VIPSeg~\cite{miao2022large} training set as our training data, and acquired the corresponding motion trajectory with Co-Tracker, as the annotation.

\subsection{Comparisons with State-of-the-Art Methods}
The generated videos are compared from four aspects: 1) Evaluation of Video Quality with FID~\cite{fid}. 2) Evaluation of Temporal Coherence with FVD~\cite{fvd}. 3) Evaluation of Object Motion with ObjMC. 4) User Study with Human Voting.

\begin{figure*}[t]
    \begin{center}
	\includegraphics[width=0.97\linewidth]{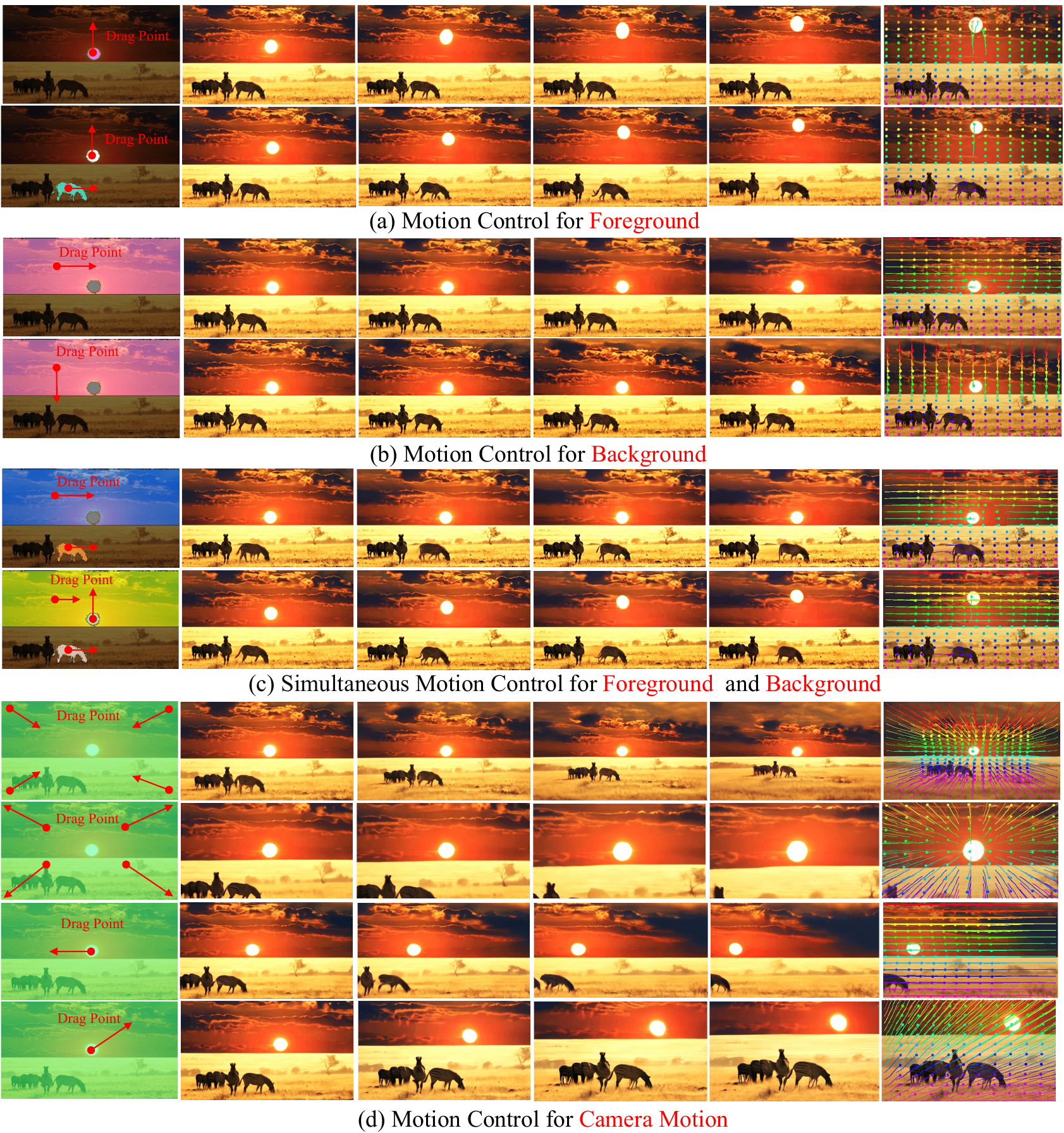}
    \end{center}
	\vspace{-0.6cm}
	\caption{\textbf{Various Motion Control from \ours.} \ours can achieve diverse motion control, such as control of foreground, background, and camera.
    }
    \vspace{-0.25cm}
\label{vis2}
\end{figure*}

\textbf{Evaluation of Video Quality on VIPSeg \texttt{val}.}
Table~\ref{VIPSeg} presents the comparison of video quality with FID on the VIPSeg \texttt{val} set. 
We control for other conditions to be the same (base architecture) and compare the performance between our method and DragNUWA.
The FID of our DragAnything reached $33.5$, significantly outperforming the current SOTA model DragNUWA with $6.3$ ($33.5$ $vs.$ $39.8$).
Figure~\ref{vis} and Figure~\ref{vis2} also demonstrate that the synthesized videos from \ours exhibit exceptionally high video quality.

\textbf{Evaluation of Temporal Coherence on VIPSeg \texttt{val}.} 
FVD~\cite{fvd} can evaluate the temporal coherence of generated videos by comparing the feature distributions in the generated video with those in the ground truth video.
We present the comparison of FVD, as shown in Table~\ref{VIPSeg}.
Compared to the performance of DragNUWA ($519.3$ FVD), our \ours achieved superior temporal coherence, \ie{} $494.8$, with a notable improvement of $24.5$.

\begin{figure*}[t]
    \begin{center}
	\includegraphics[width=0.99\linewidth]{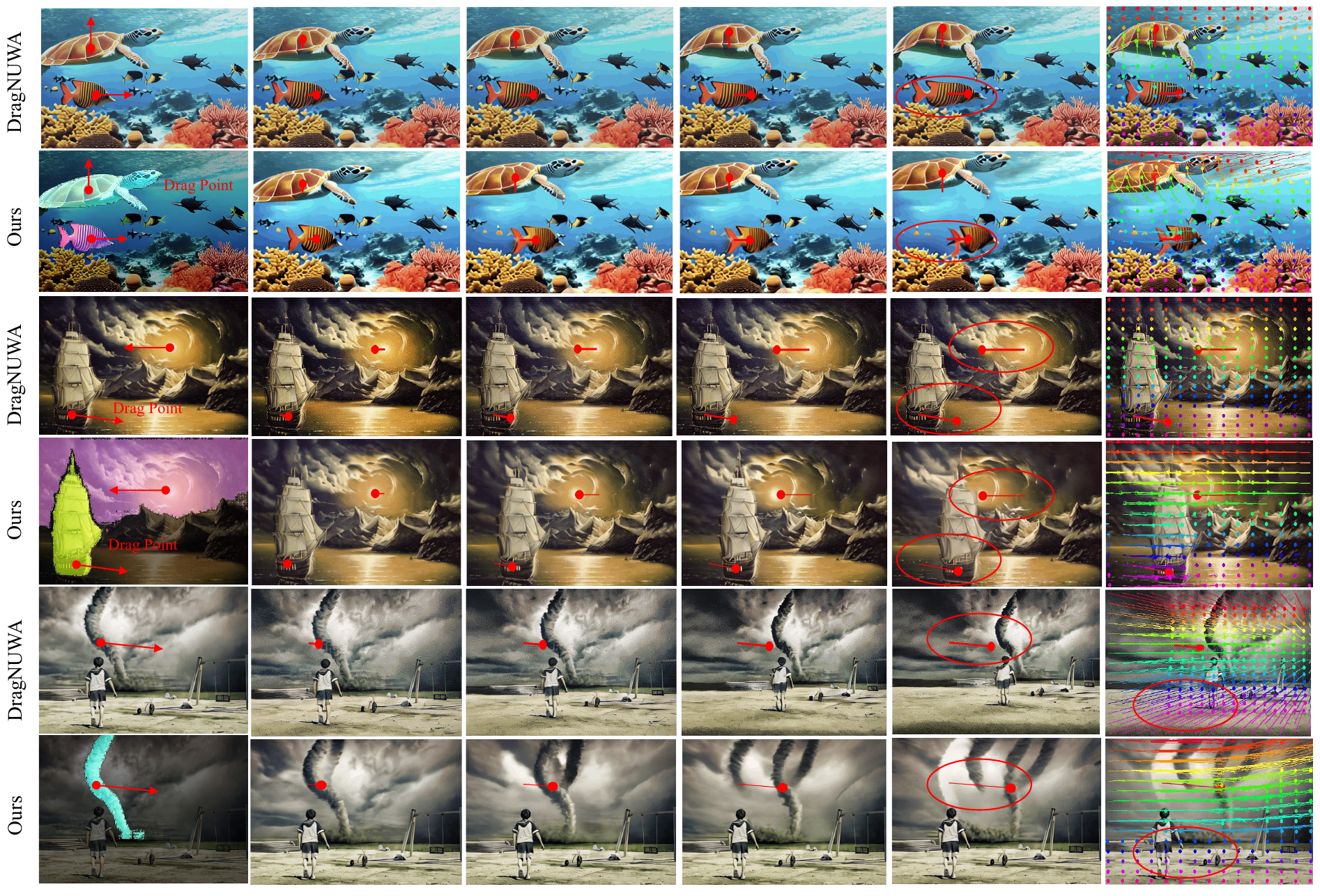}
    \end{center}
	\vspace{-0.6cm}
	\caption{\textbf{Visualization Comparison with DragNUWA.} DragNUWA leads to \textcolor{red}{distortion} of appearance (first row), \textcolor{red}{out-of-control} \texttt{sky} and \texttt{ship} (third row), \textcolor{red}{incorrect camera motion} (fifth row), while \ours enables precise control of motion. 
    }
    \vspace{-0.25cm}
\label{comparison11}
\end{figure*}

\textbf{Evaluation of Object Motion on VIPSeg \texttt{val}.}
Following MotionCtrl~\cite{wang2023motionctrl}, ObjMC is used to evaluate the motion control performance by computing the Euclidean distance between the predicted and ground truth trajectories.
Table~\ref{VIPSeg} presents the comparison of ObjMC on the VIPSeg \texttt{val} set. 
Compared to DragNUWA, our \ours achieved a new state-of-the-art performance, $305.7$, with an improvement of $18.9$.
Figure~\ref{comparison11} provides the visualization comparison between the both methods.

\begin{figure}[t]
    
    \centering
    \includegraphics[width=0.7\linewidth]{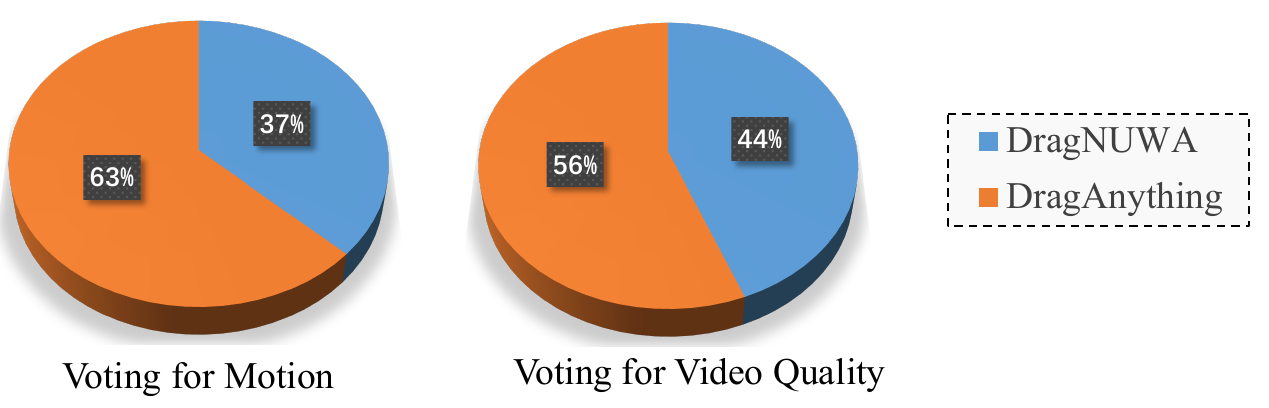}
    \vspace{-0.3cm}
    \caption{\textbf{User Study for Motion Control and Video Quality}. \ours achieved superior performance in terms of motion control and video quality.}
    \label{UserStudy}
    \vspace{-0.3cm}
\end{figure}

\textbf{User Study for Motion Control and Video Quality.}
Figure~\ref{UserStudy} presents the comparison for the user study of motion control and video quality.
Our model outperforms \ours by $26\%$ and $12\%$ in human voting for motion control and video quality, respectively.
We also provide visual comparisons in Figure~\ref{comparison11} and more visualizations in in Figure~\ref{vis}.
Our algorithm has a more accurate understanding and implementation of motion control.

\begin{table}[t]
    \centering
    \small 
    \setlength{\tabcolsep}{1mm}
    \caption{\textbf{Performance Comparison on VIPSeg \texttt{val} $256\times 256$~\cite{miao2022large}.} We only compared against DragNUWA, as other relevant works~(\eg{} Motionctrl~\cite{wang2023motionctrl}) did not release source code based on SVD~\cite{blattmann2023stable}.
    }
    \vspace{-2mm}
    \scriptsize
\begin{tabular}{l |c|ccc|c}
    Method   & Base Arch. &ObjMC$\downarrow$ & FVD$\downarrow$ &  
     FID$\downarrow$  &  
     Venue/Date
     \\
    \hline
    
    DragNUWA~\cite{yin2023dragnuwa} & SVD~\cite{blattmann2023stable} & 324.6 &  519.3 &   39.8  &  arXiv, Aug. 2023 \\
    \ours (Ours) &  SVD~\cite{blattmann2023stable} &\textbf{305.7} &  \textbf{494.8} &   \textbf{33.5}  &  - \\

    \end{tabular}
    \vspace{-5mm}
    \label{VIPSeg}
\end{table}

\subsection{Ablation Studies}

Entity representation and 2D Gaussian representation are both core components of our work.
We maintain other conditions constant and only modify the corresponding conditional embedding features.
Table~\ref{ablation11} present the ablation study for the two representations.

\textbf{Effect of Entity Representation $\bm{\mathrm{\hat{E}}}$.} 
To investigate the impact of Entity Representation $\bm{\mathrm{\hat{E}}}$, we observe the change in performance by determining whether this representation is included in the final embedding (Equation~\ref{equ2}).
As condition information $\bm{\mathrm{\hat{E}}}$ primarily affects the object motion in generating videos, we only need to compare ObjMC, while FVD and FID metrics focus on temporal consistency and overall video quality.
With Entity Representation $\bm{\mathrm{\hat{E}}}$, ObjMC of the model achieved a significant improvement($92.3$), reaching $318.4$.

\begin{table}[t]
    \centering
    \small 
    \setlength{\tabcolsep}{1mm}
    \begin{minipage}[t]{0.56\linewidth}
    \caption{\textbf{Ablation for Entity and 2D Gaussian Representation.} The combination of the both yields the greatest benefit.
    }
    \scriptsize
\begin{tabular}{ c |c |c| cc}
    Entity Rep. &  Gaussian Rep.  & ObjMC$\downarrow$ & FVD$\downarrow$ &  
     FID$\downarrow$    
     \\
    \hline
     & & 410.7 & 496.3 & 34.2 \\
    \checkmark & & 318.4 &494.5 & 34.1\\
     & \checkmark& 339.3 & 495.3& 34.0\\
     \checkmark & \checkmark & \textbf{305.7} &  494.8 &   33.5
    \end{tabular}
    \label{ablation11}
    \end{minipage}
    \hfill
   \begin{minipage}[t]{0.42\textwidth}
   \caption{\textbf{Ablation Study for Loss Mask $\bm{\mathrm{M}}$.} Loss mask can bring certain gains, especially for the ObjMC metric.
    }
    \vspace{+2mm}
    \scriptsize
\begin{tabular}{ c|c cc}
     Loss Mask $\bm{\mathrm{M}}$ &  ObjMC$\downarrow$ & FVD$\downarrow$ &  
     FID$\downarrow$    
     \\
    \hline
     &  311.1 & 500.2 & 34.3  \\
     \checkmark  & \textbf{305.7} & 494.8 & 33.5  \\
    \end{tabular}
    \label{ablation12}
    \end{minipage}
    
\end{table}

    

    

\textbf{Effect of 2D Gaussian Representation.}
Similar to Entity Representation, we observe the change in ObjMC performance by determining whether 2D Gaussian Representation is included in the final embedding.
2D Gaussian Representation resulted in an improvement of $71.4$, reaching $339.3$.
Overall, the performance is highest when both Entity and 2D Gaussian Representations are used, achieving $305.7$. 
This phenomenon suggests that the two representations have a mutually reinforcing effect.

\textbf{Effect of Loss Mask $\bm{\mathrm{M}}$.} Table~\ref{ablation12} presents the ablation for Loss Mask $\bm{\mathrm{M}}$.
When the loss mask $\bm{\mathrm{M}}$ is not used, we directly optimize the MSE loss for each pixel of the entire image.
The loss mask can bring certain gains, approximately $5.4$ of ObjMC.

\subsection{Discussion for Various Motion Control}
Our \ours is highly flexible and user-friendly, supporting diverse motion control for any entity appearing in the video. 
In this section, we will discuss the corresponding motion control, categorizing it into four types.

\textbf{Motion Control For Foreground.}
As shown in Figure~\ref{vis2} (a), 
foreground motion control is the most basic and commonly used operation.
Both the \texttt{sun} and the \texttt{horse} belong to the foreground. 
We select the corresponding region that needs to be controlled with SAM~\cite{kirillov2023segment}, and then drag any point within that region to achieve motion control over the object.
It can be observed that \ours can precisely control the movement of the sun and the horse.

\textbf{Motion Control For Background.}
Compared to the foreground, the background is usually more challenging to control because the shapes of background elements, such as \texttt{clouds}, \texttt{starry skies}, are unpredictable and difficult to characterize.
Figure~\ref{vis2} (b) demonstrates background motion control for video generation in two scenarios.
\ours can control the movement of the entire cloud layer, either to the right or further away, by dragging a point on the cloud.

\textbf{Simultaneous Motion Control for Foreground and Background.}
\ours can also simultaneously control both foreground and background, as shown in Figure~\ref{vis2} (c).
For example, by dragging three pixels, we can simultaneously achieve motion control where the \texttt{cloud layer} moves to the right, the \texttt{sun} rises upwards, and the \texttt{horse} moves to the right.
\textbf{Camera Motion Control.}
In addition to motion control for entities in the video, \ours also supports some basic control over camera motion, such as zoom in and zoom out, as shown in Figure~\ref{vis2} (d).
The user simply needs to select the entire image and then drag four points to achieve the corresponding zoom in or zoom out. Additionally, the user can also control the movement of the entire camera up, down, left, or right by dragging any point.

\section{Conclusion}
In this paper, we reevaluate the current trajectory-based motion control approach in video generation tasks and introduce two new insights: 1) Trajectory points on objects cannot adequately represent the entity. 2) For the trajectory point representation paradigm, pixels closer to the drag point exert a stronger influence, resulting in larger motions.
Addressing these two technical challenges, we present \ours, which utilizes the latent features of the diffusion model to represent each entity.
%
%
%
The proposed entity representation serves as an open-domain embedding capable of representing any object, enabling the control of motion for diverse entities, including the background. 
Extensive experiments demonstrate that our \ours achieves SOTA performance for User Study, surpassing the previous state of the art (DragNUWA) by $26\%$ in human voting.

\begin{figure*}[h]
    \begin{center}
	\includegraphics[width=0.98\linewidth]{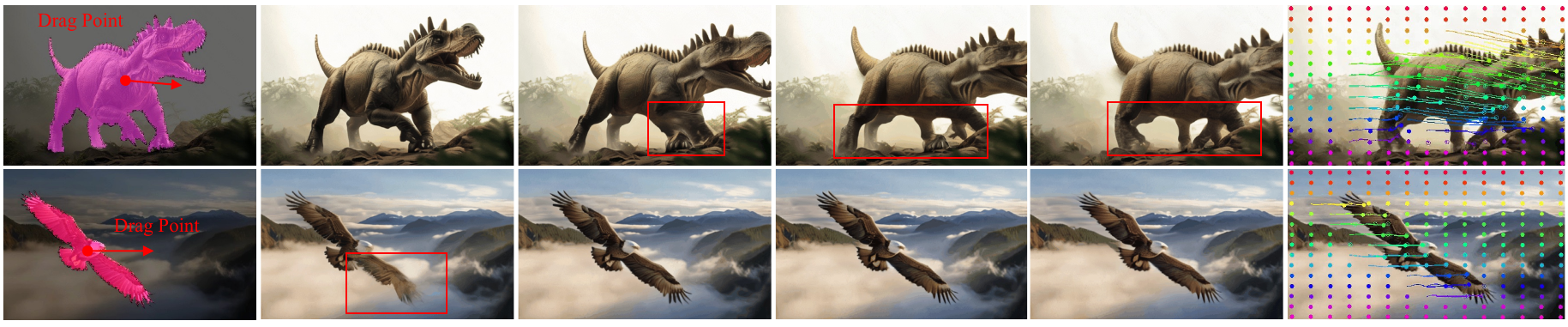}
    \end{center}
	\vspace{-0.6cm}
	\caption{\textbf{Bad Case for \ours.} \ours still has some bad cases, especially when controlling larger motions.
    }
    \vspace{-0.65cm}
\label{vis4}
\end{figure*}

\begin{figure*}[t]
    \begin{center}
	\includegraphics[width=0.98\linewidth]{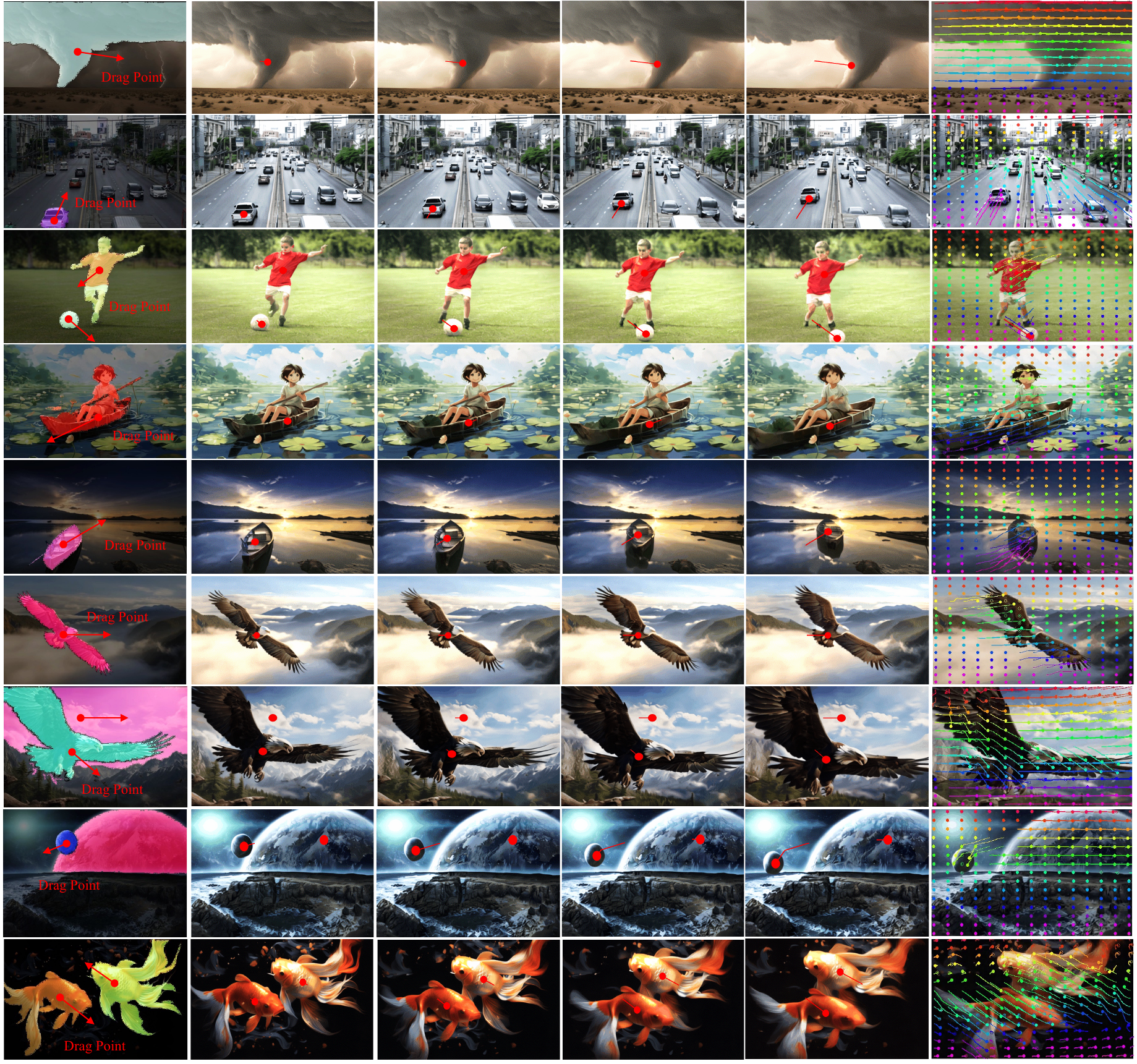}
    \end{center}
	\vspace{-0.6cm}
	\caption{\textbf{More Visualization for \ours.} 
    }
    \vspace{-0.5cm}
\label{vis3}
\end{figure*}

\section{Appendix}

\subsection{Discussion of Potential Negative Impact.}
One potential negative impact is the possibility of reinforcing biases present in the training data, as the model learns from existing datasets that may contain societal biases. 
Additionally, there is a risk of the generated content being misused, leading to the creation of misleading or inappropriate visual materials.
Furthermore, privacy concerns may arise, especially when generating videos that involve individuals without their explicit consent.
As with any other video generation technology, there is a need for vigilance and responsible implementation to mitigate these potential negative impacts and ensure ethical use.

\subsection{Limitation and Bad Case Analysis} Although our \ours has demonstrated promising performance, there are still some aspects that could be improved, which are common to current other trajectory-based video generation models: 1) Current trajectory-based motion control is limited to the 2D dimension and cannot handle motion in 3D scenes, such as controlling someone turning around or more precise body rotations.
2) Current models are constrained by the performance of the foundation model, Stable Video Diffusion~\cite{blattmann2023stable}, and cannot generate scenes with very large motions, as shown in Figure~\ref{vis4}.
It is obvious that in the first column of video frames, the legs of dinosaur don't adhere to real-world constraints. 
There are a few frames where there are five legs and some strange motions.
A similar situation occurs with the blurring of the wings of eagle in the second row.
This could be due to excessive motion, exceeding the generation capabilities of the foundation model, resulting in a collapse in video quality.
There are some potential solutions to address these two challenges.
For the first challenge, a feasible approach is to incorporate depth information into the 2D trajectory, expanding it into 3D trajectory information, thereby enabling control of object motion in 3D space.
As for the second challenge, it requires the development of a stronger foundation model to support larger and more robust motion generation capabilities.
For example, leveraging the latest text-to-video foundation from OpenAI, SORA, undoubtedly has the potential to significantly enhance the quality of generated videos.
In addition, we have provided more exquisite video cases in the supplementary materials for reference, as shown in Figure~\ref{vis3}.
For more visualizations in GIF format, please refer to \texttt{DragAnything.html} in the same directory. Simply click to open.

\bibliographystyle{splncs04}
\bibliography{arxiv}

\begin{thebibliography}{10}
\providecommand{\url}[1]{\texttt{#1}}
\providecommand{\urlprefix}{URL }
\providecommand{\doi}[1]{https://doi.org/#1}

\bibitem{pikalab}
https://www.pika.art/

\bibitem{ardino2021click}
Ardino, P., De~Nadai, M., Lepri, B., Ricci, E., Lathuili{\`e}re, S.: Click to move: Controlling video generation with sparse motion. In: Proceedings of the IEEE/CVF International Conference on Computer Vision. pp. 14749--14758 (2021)

\bibitem{blattmann2023stable}
Blattmann, A., Dockhorn, T., Kulal, S., Mendelevitch, D., Kilian, M., Lorenz, D., Levi, Y., English, Z., Voleti, V., Letts, A., et~al.: Stable video diffusion: Scaling latent video diffusion models to large datasets. arXiv preprint arXiv:2311.15127  (2023)

\bibitem{blattmann2021ipoke}
Blattmann, A., Milbich, T., Dorkenwald, M., Ommer, B.: ipoke: Poking a still image for controlled stochastic video synthesis. In: Proceedings of the IEEE/CVF International Conference on Computer Vision. pp. 14707--14717 (2021)

\bibitem{blattmann2021understanding}
Blattmann, A., Milbich, T., Dorkenwald, M., Ommer, B.: Understanding object dynamics for interactive image-to-video synthesis. In: Proceedings of the IEEE/CVF Conference on Computer Vision and Pattern Recognition. pp. 5171--5181 (2021)

\bibitem{blattmann2023align}
Blattmann, A., Rombach, R., Ling, H., Dockhorn, T., Kim, S.W., Fidler, S., Kreis, K.: Align your latents: High-resolution video synthesis with latent diffusion models. In: Proceedings of the IEEE/CVF Conference on Computer Vision and Pattern Recognition. pp. 22563--22575 (2023)

\bibitem{cao2017realtime}
Cao, Z., Simon, T., Wei, S.E., Sheikh, Y.: Realtime multi-person 2d pose estimation using part affinity fields. In: Proceedings of the IEEE conference on computer vision and pattern recognition. pp. 7291--7299 (2017)

\bibitem{chen2023videocrafter1}
Chen, H., Xia, M., He, Y., Zhang, Y., Cun, X., Yang, S., Xing, J., Liu, Y., Chen, Q., Wang, X., et~al.: Videocrafter1: Open diffusion models for high-quality video generation. arXiv preprint arXiv:2310.19512  (2023)

\bibitem{chen2023motion}
Chen, T.S., Lin, C.H., Tseng, H.Y., Lin, T.Y., Yang, M.H.: Motion-conditioned diffusion model for controllable video synthesis. arXiv preprint arXiv:2304.14404  (2023)

\bibitem{chen2023control}
Chen, W., Wu, J., Xie, P., Wu, H., Li, J., Xia, X., Xiao, X., Lin, L.: Control-a-video: Controllable text-to-video generation with diffusion models. arXiv preprint arXiv:2305.13840  (2023)

\bibitem{chen2023anydoor}
Chen, X., Huang, L., Liu, Y., Shen, Y., Zhao, D., Zhao, H.: Anydoor: Zero-shot object-level image customization. arXiv preprint arXiv:2307.09481  (2023)

\bibitem{dai2023emu}
Dai, X., Hou, J., Ma, C.Y., Tsai, S., Wang, J., Wang, R., Zhang, P., Vandenhende, S., Wang, X., Dubey, A., et~al.: Emu: Enhancing image generation models using photogenic needles in a haystack. arXiv preprint arXiv:2309.15807  (2023)

\bibitem{esser2023structure}
Esser, P., Chiu, J., Atighehchian, P., Granskog, J., Germanidis, A.: Structure and content-guided video synthesis with diffusion models. In: Proceedings of the IEEE/CVF International Conference on Computer Vision. pp. 7346--7356 (2023)

\bibitem{girdhar2023emu}
Girdhar, R., Singh, M., Brown, A., Duval, Q., Azadi, S., Rambhatla, S.S., Shah, A., Yin, X., Parikh, D., Misra, I.: Emu video: Factorizing text-to-video generation by explicit image conditioning. arXiv preprint arXiv:2311.10709  (2023)

\bibitem{gu2024mix}
Gu, Y., Wang, X., Wu, J.Z., Shi, Y., Chen, Y., Fan, Z., Xiao, W., Zhao, R., Chang, S., Wu, W., et~al.: Mix-of-show: Decentralized low-rank adaptation for multi-concept customization of diffusion models. Advances in Neural Information Processing Systems  \textbf{36} (2024)

\bibitem{gu2023videoswap}
Gu, Y., Zhou, Y., Wu, B., Yu, L., Liu, J.W., Zhao, R., Wu, J.Z., Zhang, D.J., Shou, M.Z., Tang, K.: Videoswap: Customized video subject swapping with interactive semantic point correspondence. arXiv preprint arXiv:2312.02087  (2023)

\bibitem{guo2023sparsectrl}
Guo, Y., Yang, C., Rao, A., Agrawala, M., Lin, D., Dai, B.: Sparsectrl: Adding sparse controls to text-to-video diffusion models. arXiv preprint arXiv:2311.16933  (2023)

\bibitem{guo2023animatediff}
Guo, Y., Yang, C., Rao, A., Wang, Y., Qiao, Y., Lin, D., Dai, B.: Animatediff: Animate your personalized text-to-image diffusion models without specific tuning. arXiv preprint arXiv:2307.04725  (2023)

\bibitem{hao2018controllable}
Hao, Z., Huang, X., Belongie, S.: Controllable video generation with sparse trajectories. In: Proceedings of the IEEE Conference on Computer Vision and Pattern Recognition. pp. 7854--7863 (2018)

\bibitem{he2023efficientdm}
He, Y., Liu, J., Wu, W., Zhou, H., Zhuang, B.: Efficientdm: Efficient quantization-aware fine-tuning of low-bit diffusion models. arXiv preprint arXiv:2310.03270  (2023)

\bibitem{he2024ptqd}
He, Y., Liu, L., Liu, J., Wu, W., Zhou, H., Zhuang, B.: Ptqd: Accurate post-training quantization for diffusion models. Advances in Neural Information Processing Systems  \textbf{36} (2024)

\bibitem{ho2022imagen}
Ho, J., Chan, W., Saharia, C., Whang, J., Gao, R., Gritsenko, A., Kingma, D.P., Poole, B., Norouzi, M., Fleet, D.J., et~al.: Imagen video: High definition video generation with diffusion models. arXiv preprint arXiv:2210.02303  (2022)

\bibitem{ho2020denoising}
Ho, J., Jain, A., Abbeel, P.: Denoising diffusion probabilistic models. Advances in Neural Information Processing Systems  \textbf{33},  6840--6851 (2020)

\bibitem{ho2022video}
Ho, J., Salimans, T., Gritsenko, A., Chan, W., Norouzi, M., Fleet, D.J.: Video diffusion models. arXiv:2204.03458  (2022)

\bibitem{karaev2023cotracker}
Karaev, N., Rocco, I., Graham, B., Neverova, N., Vedaldi, A., Rupprecht, C.: Cotracker: It is better to track together. arXiv:2307.07635  (2023)

\bibitem{kirillov2023segment}
Kirillov, A., Mintun, E., Ravi, N., Mao, H., Rolland, C., Gustafson, L., Xiao, T., Whitehead, S., Berg, A.C., Lo, W.Y., et~al.: Segment anything. arXiv preprint arXiv:2304.02643  (2023)

\bibitem{loshchilov2017decoupled}
Loshchilov, I., Hutter, F.: Decoupled weight decay regularization. arXiv preprint arXiv:1711.05101  (2017)

\bibitem{ma2023trailblazer}
Ma, W.D.K., Lewis, J., Kleijn, W.B.: Trailblazer: Trajectory control for diffusion-based video generation. arXiv preprint arXiv:2401.00896  (2023)

\bibitem{ma2023follow}
Ma, Y., He, Y., Cun, X., Wang, X., Shan, Y., Li, X., Chen, Q.: Follow your pose: Pose-guided text-to-video generation using pose-free videos. arXiv preprint arXiv:2304.01186  (2023)

\bibitem{miao2022large}
Miao, J., Wang, X., Wu, Y., Li, W., Zhang, X., Wei, Y., Yang, Y.: Large-scale video panoptic segmentation in the wild: A benchmark. In: Proceedings of the IEEE/CVF Conference on Computer Vision and Pattern Recognition. pp. 21033--21043 (2022)

\bibitem{oquab2023dinov2}
Oquab, M., Darcet, T., Moutakanni, T., Vo, H., Szafraniec, M., Khalidov, V., Fernandez, P., Haziza, D., Massa, F., El-Nouby, A., et~al.: Dinov2: Learning robust visual features without supervision. arXiv preprint arXiv:2304.07193  (2023)

\bibitem{ramesh2022hierarchical}
Ramesh, A., Dhariwal, P., Nichol, A., Chu, C., Chen, M.: Hierarchical text-conditional image generation with clip latents. arXiv preprint arXiv:2204.06125  \textbf{1}(2), ~3 (2022)

\bibitem{rombach2022high}
Rombach, R., Blattmann, A., Lorenz, D., Esser, P., Ommer, B.: High-resolution image synthesis with latent diffusion models. In: Proceedings of the IEEE/CVF conference on computer vision and pattern recognition. pp. 10684--10695 (2022)

\bibitem{unet}
Ronneberger, O., Fischer, P., Brox, T.: U-net: Convolutional networks for biomedical image segmentation. In: MICCAI (2015)

\bibitem{saharia2022photorealistic}
Saharia, C., Chan, W., Saxena, S., Li, L., Whang, J., Denton, E.L., Ghasemipour, K., Gontijo~Lopes, R., Karagol~Ayan, B., Salimans, T., et~al.: Photorealistic text-to-image diffusion models with deep language understanding. Advances in Neural Information Processing Systems  \textbf{35},  36479--36494 (2022)

\bibitem{fid}
Seitzer, M.: {pytorch-fid: FID Score for PyTorch}. \url{https://github.com/mseitzer/pytorch-fid} (2020)

\bibitem{tang2024emergent}
Tang, L., Jia, M., Wang, Q., Phoo, C.P., Hariharan, B.: Emergent correspondence from image diffusion. Advances in Neural Information Processing Systems  \textbf{36} (2024)

\bibitem{SORA}
Tim, B., Peebles, B., Holmes, C., DePue, W., Guo, Y., Jing, L., Schnurr, D., Taylor, J., Troy, L., Luhman, E., Ng, C.W.Y., Wang, R., Ramesh, A.: Video generation models as world simulators (2024)

\bibitem{fvd}
Unterthiner, T., Van~Steenkiste, S., Kurach, K., Marinier, R., Michalski, M., Gelly, S.: Towards accurate generative models of video: A new metric \& challenges. arXiv preprint arXiv:1812.01717  (2018)

\bibitem{wang2023seggpt}
Wang, X., Zhang, X., Cao, Y., Wang, W., Shen, C., Huang, T.: Seggpt: Segmenting everything in context. arXiv preprint arXiv:2304.03284  (2023)

\bibitem{wang2023lavie}
Wang, Y., Chen, X., Ma, X., Zhou, S., Huang, Z., Wang, Y., Yang, C., He, Y., Yu, J., Yang, P., et~al.: Lavie: High-quality video generation with cascaded latent diffusion models. arXiv preprint arXiv:2309.15103  (2023)

\bibitem{wang2023motionctrl}
Wang, Z., Yuan, Z., Wang, X., Chen, T., Xia, M., Luo, P., Shan, Y.: Motionctrl: A unified and flexible motion controller for video generation. arXiv preprint arXiv:2312.03641  (2023)

\bibitem{wu2023tune}
Wu, J.Z., Ge, Y., Wang, X., Lei, S.W., Gu, Y., Shi, Y., Hsu, W., Shan, Y., Qie, X., Shou, M.Z.: Tune-a-video: One-shot tuning of image diffusion models for text-to-video generation. In: Proceedings of the IEEE/CVF International Conference on Computer Vision. pp. 7623--7633 (2023)

\bibitem{wu2023paragraph}
Wu, W., Li, Z., He, Y., Shou, M.Z., Shen, C., Cheng, L., Li, Y., Gao, T., Zhang, D., Wang, Z.: Paragraph-to-image generation with information-enriched diffusion model. arXiv preprint arXiv:2311.14284  (2023)

\bibitem{wu2024datasetdm}
Wu, W., Zhao, Y., Chen, H., Gu, Y., Zhao, R., He, Y., Zhou, H., Shou, M.Z., Shen, C.: Datasetdm: Synthesizing data with perception annotations using diffusion models. Advances in Neural Information Processing Systems  \textbf{36} (2024)

\bibitem{wu2023diffumask}
Wu, W., Zhao, Y., Shou, M.Z., Zhou, H., Shen, C.: Diffumask: Synthesizing images with pixel-level annotations for semantic segmentation using diffusion models. In: Proceedings of the IEEE/CVF International Conference on Computer Vision. pp. 1206--1217 (2023)

\bibitem{xing2023survey}
Xing, Z., Feng, Q., Chen, H., Dai, Q., Hu, H., Xu, H., Wu, Z., Jiang, Y.G.: A survey on video diffusion models. arXiv preprint arXiv:2310.10647  (2023)

\bibitem{xue2023raphael}
Xue, Z., Song, G., Guo, Q., Liu, B., Zong, Z., Liu, Y., Luo, P.: Raphael: Text-to-image generation via large mixture of diffusion paths. arXiv preprint arXiv:2305.18295  (2023)

\bibitem{yin2023dragnuwa}
Yin, S., Wu, C., Liang, J., Shi, J., Li, H., Ming, G., Duan, N.: Dragnuwa: Fine-grained control in video generation by integrating text, image, and trajectory. arXiv preprint arXiv:2308.08089  (2023)

\bibitem{zhang2024moonshot}
Zhang, D.J., Li, D., Le, H., Shou, M.Z., Xiong, C., Sahoo, D.: Moonshot: Towards controllable video generation and editing with multimodal conditions. arXiv preprint arXiv:2401.01827  (2024)

\bibitem{zhang2023show}
Zhang, D.J., Wu, J.Z., Liu, J.W., Zhao, R., Ran, L., Gu, Y., Gao, D., Shou, M.Z.: Show-1: Marrying pixel and latent diffusion models for text-to-video generation. arXiv preprint arXiv:2309.15818  (2023)

\bibitem{zhang2023adding}
Zhang, L., Rao, A., Agrawala, M.: Adding conditional control to text-to-image diffusion models. In: Proceedings of the IEEE/CVF International Conference on Computer Vision. pp. 3836--3847 (2023)

\bibitem{zhang2023i2vgen}
Zhang, S., Wang, J., Zhang, Y., Zhao, K., Yuan, H., Qin, Z., Wang, X., Zhao, D., Zhou, J.: I2vgen-xl: High-quality image-to-video synthesis via cascaded diffusion models. arXiv preprint arXiv:2311.04145  (2023)

\bibitem{zhang2023controlvideo}
Zhang, Y., Wei, Y., Jiang, D., Zhang, X., Zuo, W., Tian, Q.: Controlvideo: Training-free controllable text-to-video generation. arXiv preprint arXiv:2305.13077  (2023)

\bibitem{zhao2023motiondirector}
Zhao, R., Gu, Y., Wu, J.Z., Zhang, D.J., Liu, J., Wu, W., Keppo, J., Shou, M.Z.: Motiondirector: Motion customization of text-to-video diffusion models. arXiv preprint arXiv:2310.08465  (2023)

\bibitem{zhou2022magicvideo}
Zhou, D., Wang, W., Yan, H., Lv, W., Zhu, Y., Feng, J.: Magicvideo: Efficient video generation with latent diffusion models. arXiv preprint arXiv:2211.11018  (2022)

\end{thebibliography}
\end{document}